\pdfoutput=1
\documentclass[11pt]{article}

\usepackage{acl}

\usepackage{times}
\usepackage{latexsym}

\usepackage{amsmath,amsfonts,bm}

\def\eqref#1{equation~\ref{#1}}
\def\1{\bm{1}}

\DeclareMathAlphabet{\mathsfit}{\encodingdefault}{\sfdefault}{m}{sl}
\SetMathAlphabet{\mathsfit}{bold}{\encodingdefault}{\sfdefault}{bx}{n}

\usepackage{array}
\usepackage{float}
\usepackage[lofdepth,lotdepth]{subfig}
\usepackage{graphicx}
\usepackage{makecell}
\usepackage{url}
\usepackage{hhline}
\usepackage{hyperref}
\usepackage{amssymb}
\usepackage{amsmath}
\usepackage{booktabs}

\usepackage[T1]{fontenc}
\usepackage[utf8]{inputenc}

\usepackage{microtype}

\title{Clean or Annotate: \\ How to Spend a Limited Data Collection Budget}

\author{Derek Chen \\
ASAPP, New York, NY \\
\texttt{dchen@asapp.com} \\
\And
Zhou Yu \\
Columbia University, NY \\
\texttt{zy2461@columbia.edu} \\
\And
Samuel R. Bowman \\
New York University, NY \\
\texttt{bowman@nyu.edu}
}

\begin{document}
\maketitle
\begin{abstract}
% Supervised learning with deep neural networks requires a massive amount of labeled examples to achieve impressive performance. 
%  This method is still interesting even if you need only moderate numbers of labels, or if you're getting labels to fix long-tail performance issues, rather than get a model up to basically-good performance
Crowdsourcing platforms are often used to collect datasets for training machine learning models, despite % their % but come with
higher levels of inaccurate labeling compared to expert labeling.
There are two common strategies to manage the impact of such noise: The first 
% A simple solution for improving reliability   % eliciting and aggregating multiple annotations per example
involves aggregating redundant annotations, but comes at the expense of labeling substantially fewer examples. % for a given budget.  
Secondly, prior works have also considered using the entire annotation budget to label as many examples as possible and subsequently apply denoising algorithms to implicitly clean the dataset.
% However, these methods often make assumptions about the noise distribution, which may not necessarily hold in real world scenarios.  % For example, they assume that the noise comes from deliberate spammers. 
We find a middle ground and propose an approach which reserves a fraction of annotations to \textit{explicitly} clean up % , offenders
highly probable error samples to optimize the annotation process. In particular, we allocate a large portion of the labeling budget to form an initial dataset used to train a model.  This model is then used to identify specific examples that appear most likely to be incorrect, which we spend the remaining budget to relabel.  % and train a new model with the cleaned data, allowing us to gather many examples while still maintaining high label quality.  % conducted Extensive
Experiments across three model variations and four natural language processing tasks show our approach outperforms or matches both label aggregation and advanced denoising methods designed to handle noisy labels when allocated the same finite annotation budget.
% Compared to methods designed to handle noisy labels, our approach consistently surpasses their performance across numerous model variations and natural language processing tasks when allocated the same annotation budget.
% Extensive experiments across a number of model variations and natural language processing tasks confirm our approach's benefits.  % our approach's power and simplicity.
% \input{sections/0_abstract.tex}
\end{abstract}

\section{Introduction}
% 1) ML often relies on data labeling, and cost and speed considerations generally lead people to use non-specialist crowdwork, often through platforms like MTurk. This means a substantial error rate. This paper looks at ways to reduce that error rate in training data.
% considerable, enormous, huge, other considerations
% Goal is to train deep neural networks with crowdsourced data.  At reasonable costs.

% Training large deep learning models in m
Modern machine learning often depends on heavy data annotation efforts.
% However, relying on expert annotators to scale can quickly surpass any data collection budget.
To keep costs in check while maintaining speed and scalability, many people turn to non-specialist crowd-workers through platforms like Mechanical Turk.  
Although crowdsourcing reduces costs to a reasonable level, it also tends to produce substantially higher error rates compared with expert labeling.
% This paper explores practical methods for reducing the error rate in training data while operating with a fixed annotation budget.
% 2) Classic approach: Multiple annotation. Common (cite actual use cases, not just studies about the method), but much more costly.
The classic approach for improving reliability in classification tasks %in the face of noisy labels,  per example
is to perform redundant annotations which are later aggregated using a majority vote to form a single gold label \citep{snow08nlp, sap2019atomic, potts20sentiment, sap2019socialiqa}.
This solution is easy to understand and implement, but comes at the expense of severely reducing the number of labeled examples available for training. %  Additionally, majority voting can cause biased results when the task is difficult since assumptions that the labels and the data are independent conditioned on the ground truth become inaccurate \citep{cao19maxmig}.
% assumptions about the expert labeling noise is mutually independent is untrue
% assumptions about mutual independence in label noise

\begin{figure}
  \includegraphics[width=\linewidth]{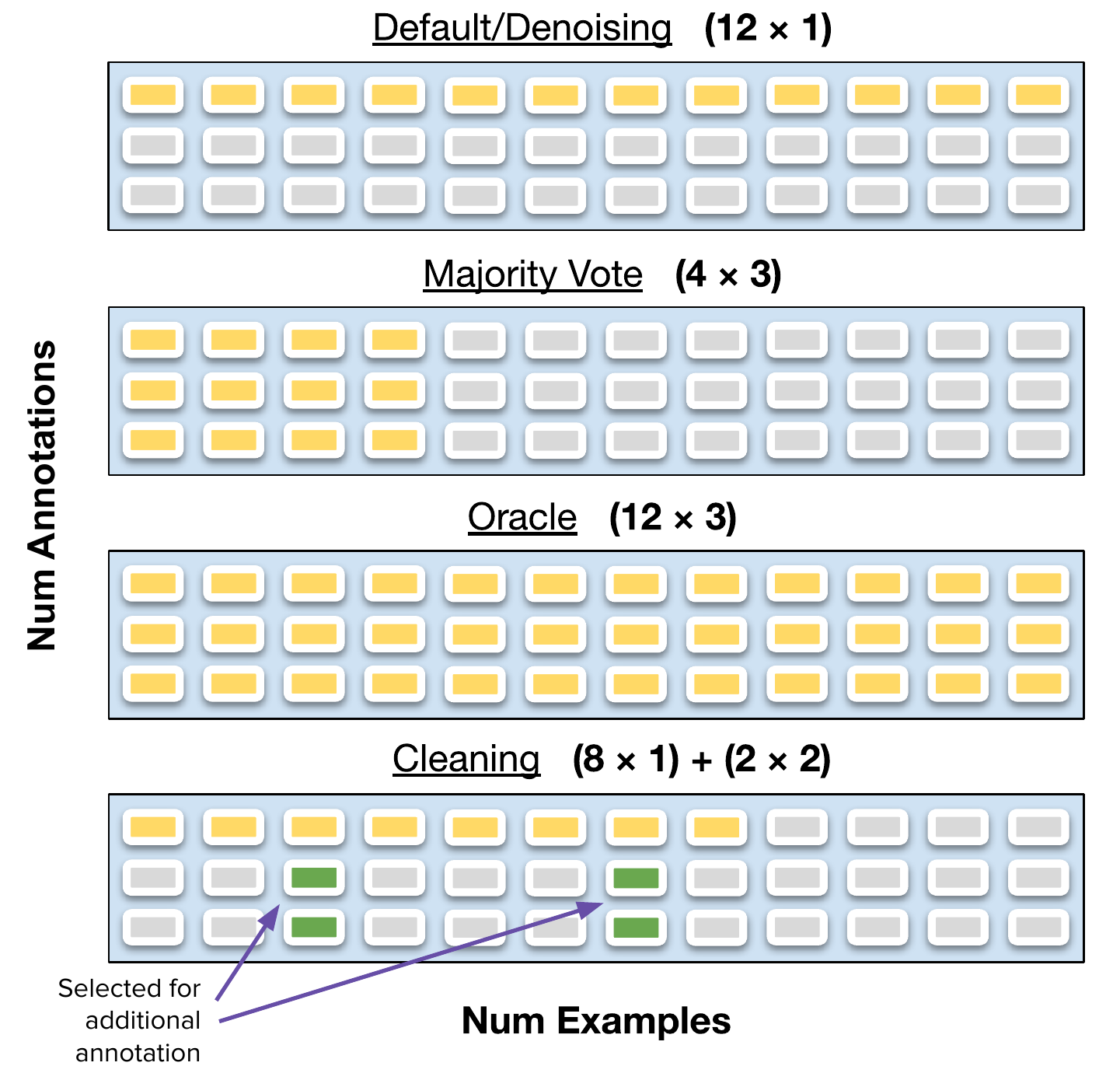}
  \caption{Data cleaning reserves a small portion of the annotation budget for targeted relabeling of examples that are identified as especially likely to be noisy. In contrast, the default and denoising methods spend the entire budget upfront, yielding lower quality data.}
  \label{fig:front_page}
\end{figure}

% 3) Newer approach: Automatic cleaning/reweighting. More complex. 

% To address variation in worker skill, existing methods tackling this problem either try to identify and correct the wrong labels or reweigh the data terms in the loss function according to the inferred noisy rates.
As an alternative, researchers have made great strides in designing automatic label cleaning methods, noise-insensitive training schemes and other mechanisms to work with noisy data \citep{ sukhbaatar2014noisycnn, han18coteaching, tanaka18joint}. % and other ways to deal with noise. and other methods of label aggregation. % depend on shaky foundations to work. 
For example, some methods learn a noise transition matrix for reweighting the label \citep{dawid1979maximum, goldberger17adaptation}, while others modify the loss \citep{ghosh2017mae, patrini17forward}.  Another set of options generate cleaned examples from mislabeled ones through semi-supervised pseudo-labeling \citep{jiang18mentornet, li2020dividemix}. %   but assume that label noise is derived from a single source which can be identified and reversed.   There are 
However, empirically getting many of these techniques to work well in practice is often a struggle due to the difficulty of training extra model components.    % They work by modeling the noise at the class level, annotator level or example level.   these make many assumptions about the source of noise, which may be false.  The source of issue may not be the label confusion, annotator spamming or example noise. but rather the ambiguity of the underlying data
% Others , but carry the assumption that biased examples can generate unbiased ones. %  These make assumptions that a model biased by noisy data is still able to produce new samples that are unbiased.  
% As a summary, note that majority vote uses annotations to gain quality, but sacrifices the number of examples.  whereas advanced methods use annotations to gain quantity, but sacrifices the potential quality and simplicity.  we see these two ends as being complementary extremes and suggest something in the middle.  Roughly speaking, we propose data cleaning through targeted relabeling that uses most annotations to gain quantity (like the advanced methods), but also saves a small portion for later to also gain the quality (like the redundant labeling techniques).

% 4) We propose a hybrid that we think can do better: Sketch the method here.
We avoid the complexity of %, critical flaws that come with   , we sidestep the problem
repairing or reweighting the labels of existing annotations by instead obtaining wholly new annotations from crowdworkers for a selected subset of samples.  In doing so, our proposed methods require no extra model parameters to train, yet still retains the benefits of high label quality.  %  from denoising methods we well as the high quality benefits of label aggregation.  
Concretely, we start by allocating a large portion of the labeling budget to obtain an initial training dataset. The examples in this dataset are annotated in a single pass, and we would expect some percentage of them to be incorrectly labeled.  However, enough of the labels should be correct to train a reasonable base model.  Next, we %consider several methods that 
take advantage of the recently trained model to identify incorrectly labeled examples, and  % specific examples that appear most likely to be incorrect. 
then spend the remaining budget to relabel those examples.
Finally, we train a new model using the original data combined with the 
cleaned data.
% computational time

The key ingredient of our method is a function for selecting which examples to re-annotate.  We consider multiple approaches for identifying candidates for relabeling, none of which have been applied before to denoising data within NLP settings. % including how the examples performed during training, which examples the model gets wrong and how examples compare to each other when encoded into a shared latent vector space. 
In all cases, relabeling the target examples relies on neither training any extra model components nor on tuning sensitive hyper-parameters. By using the existing annotation pipeline, the implementation becomes relatively trivial.
% Finally, unlike pseudo-labeling, our cleaning uses humans-in-the-loop to maintain high quality.

% 6) We evaluate this using real annotations on XYZ models/tasks. We find XYZ impressive summary statistics, and observe XYZ side observations. We recommend XYZ as a cost-effective default technique for any crowdsourced training-data annotation effort. 
To test the generalizability of our method, we compare against multiple baselines on four tasks spanning multiple natural language formats.  This departs from previous studies on human labeling in NLP, which focus exclusively on text classification \citep{wang19lwnl, jindal2019effective, tayal20classify}.  The control baseline and denoising baselines perform a single annotation per example. The majority vote baseline triples the annotations per example, but consequently is trained on only one third the number of examples to meet the annotation budget.  We lastly include an oracle baseline that lifts the restriction on a fixed budget and instead uses all available annotations. %  different NLP formats. This is generalizable to many NLP tasks, such as span selection and multiple-choice selection, which is different from prior work only applied to CV.    % state-of-the-art transformer models
We test across three model types, ranging from small ones taking minutes to train up to large transformer models which require a week to reach convergence. We find that under the same fixed annotation budget, cleaning methods match or surpass all baselines. % Extensive experiments across multiple datasets and model types show that relabeling data often has similar or even superior performance with much less code complexity.

% 7) Summary of contributions
In summary, our contributions include: 
\begin{enumerate}
    \item We examine an alternative direction to learning with noisy labels that appear when data is collected under low-resource settings.
% label cleaning applicable to real-world settings that does not make  assumptions about the source of the noise or the distribution of noise.
    \item We build four versions of our approach that vary in how they target examples to relabel.
    \item We compare against a number of baselines, many of which have never been implemented before in the natural language setting.
    % \item We conduct analysis to understand the strengths and weaknesses of our approaches
\end{enumerate}
Overall, our \textit{Large Loss} method, which selects examples for relabeling by the size of their training loss, 
performs the best out of all variations we consider despite requiring no extra parameters to train. %  We plan to release all related code and data.
%\footnote{ \href{https://github.com/asappresearch/clean}{\texttt{https://github.com/asappresearch/clean}}}
\section{Related Work}
\label{related}
The standard %default,  corrupted labels, severely
method for learning in the presence of unreliable annotation is to perform redundant annotation, where each example is annotated multiple times and a simple majority vote determines the final label \citep{snow2004hypernym, russakovsky2015imagenet, bowman2015snli}.  While effective, this can be costly since it severely reduces the amount of data collected. To tackle this problem, researchers have developed several alternative methods for dealing with noisy data that can be broken down into three categories.

\paragraph{Denoising Techniques}
Noisy training examples can be thought of as the result of perturbing the true, underlying labels by some source of noise.  
One group of methods assume the source of noise is from confusing one label \textit{class} for another, and is resolved by reverting the errors through a noise transition matrix \citep{sukhbaatar2014noisycnn, goldberger17adaptation}.
%  mixed up 
%   Symmetric noise - randomly noisy, labels can uniformly flip into other classes
%   Asymmetric noise (aka. pair-flipping) - certain classes are particularly confusing
%   Irregular asymmetric noise - certain classes get messed up more often than others, but there are no obvious patterns to exploit.  This is commonly the case in real-life datasets.
Other methods work under the assumption that labeling errors occur due to \textit{annotator} biases \citep{raykar2009whomtotrust, rodrigues18crowdlayer}, such as non-expert labelers \citep{welinder10crowds, guan2018whosaid} or spammers \citep{hovy2013mace, khetan18mbem}. % Annotators may lack relevant expertise  or engage in spamming behavior , leading to incorrect labels. 
Finally, some methods model the noise of each individual \textit{example}, either through expectation-maximization \citep{dawid1979maximum, whitehill2009whosevote, mnih2012aerial}, or neural networks \citep{felt2016semantic, jindal2019effective}.
% hoping to minimize the importance of noisy instances while boosting the importance of clean ones.

% \paragraph{Loss Modification}
Another set of methods modify the loss function to make the model more robust to noise  \citep{patrini17forward}.  For example, some methods add a regularization term \citep{tanno2019regularize}, while others bound the amount of loss contributed by individual training examples \citep{ghosh2017mae, zhang2018generalized}.
The learning procedure can also be modified such that the importance of training examples is dynamically reweighted to prevent overfitting to noise \citep{jiang18mentornet}.  
% \paragraph{Pseudo-Labeling}

Pseudo-labeling represents a final set of methods that either devise new labels for noisy data \citep{reed14bootstrap, tanaka18joint} or generate wholly new training examples \citep{arazo2019unsupervised, li2020dividemix}. Other approaches from this family use two distinct networks to produce examples for each other to learn from \citep{han18coteaching, yu2019coteachplus}.

\paragraph{Budget Constrained Data Collection}
Our work also falls under research studying how to maximize the benefit of labeled data given a fixed annotation budget. % find that model-based EM can be quite powerful in modeling annotator noise
\citet{khetan16achieve} apply model-based EM to model annotator noise, allowing singly-labeled data to outperform multiply-labeled data when annotation quality goes above a certain threshold.  \citet{bai2021pretrain} show that similar trade-offs exist when performing domain adaptation on a constrained budget. \citet{zhang21learning} observe that difficult examples benefit from additional annotations, so optimal spending actually varies the amount of labels given to each example. % of a finite budget
Our approach actively targets examples for relabeling based on its likelihood of noise, %based on its training performance,  from their work since 
whereas they randomly select examples for multi-labeling without considering its characteristics.  % Lastly, we note that all the above works focus exclusively on computer vision, with only a few exceptions targeting natural language processing \citep{snow08nlp, wang19lwnl}.

\paragraph{Human in the Loop} Finally, our work is also related to data labeling with humans.  Annotators can be assisted through iterative labeling where models suggest labels for each training example \citep{settles2011closing, schulz19analysis}, or through active learning where models suggest which examples to label \citep{settles2008active, ash20badge}.  In both cases, forward facing decisions are made on incoming batches of \textit{unlabeled} data.  In contrast, our methods look back to previously collected data to select examples for \textit{relabeling}.  These activities are orthogonal to each other and can both be included when training a model. (See Appendix~\ref{sec:scheme})

Lastly, re-active learning from \cite{sheng08getanother, lin2016reactive} proposes to relabel examples based on their predicted impact by retraining a classifier from scratch for every iteration of annotation. Accordingly, their method is impractical when adapted to the large Transformer models studied in this paper\footnote{Training a large language model (such as RoBERTa-Large) until convergence can easily take a day or longer.  Doing so each time for 12k annotations would take 30+ years.}.  Instead, we identify examples to relabel through much less computationally expensive means, making the process tractable for real-life deployment.

\section{Methods Under Study}
We study how to maximize model performance given a static data annotation budget. Concretely, we are given some model $M$ for a target task, along with a budget as measured by $B$ number of annotations, where each annotation allows us to apply a possibly noisy labeling function $f_r(x)$, where $r$ is the number of redundant annotations applied to a single example.  Annotating some set of unlabeled instances produces noisy examples $(X, f_r(X)) = (X, \Tilde{Y})$.  Our goal is to achieve the best score possible for some primary evaluation metric $S$ on a given task by cleaning the noisy labels $\Tilde{Y} \xrightarrow{clean} Y$.  Afterwards, we train a model with the cleaned data and then test it on a separate test set. For all our experiments, we set $B = 12,000$ as the total annotation budget.

As a default setting, we start with a \textit{Control} baseline which uses the entire budget to annotate 12k examples, once each $(n=12,000; r=1)$. To simulate a single annotation, we randomly sample a label from the set of labels offered for each example by the dataset.  To obtain more accurate labels, people often perform multiple annotations on each example and use \textit{Majority Vote} to aggregate the annotations.  Accordingly, as a second baseline we annotate 4k examples three times each $(n=4,000; r=3)$, matching the same total budget as before.  In the event of a tie, we randomly select one of the candidate labels.  Finally, we also include an \textit{Oracle} baseline which uses the gold label for 12k examples $(n=12,000; r=3|5)$. The gold label is either given by the dataset or generated by majority vote, where the label might result from aggregating five annotations rather than just three annotations.

\subsection{Noise Correction Baselines} 
We consider four advanced baselines, all of which perform a single annotation per example $(n=12,000, r=1)$ as seen in Figure \ref{fig:front_page}.
(1) \cite{goldberger17adaptation} propose applying a noise \textit{Adaptation} layer which models the error probability of label classes. This layer is initialized as an identity matrix, which biases the layer to act as if there is no confusion in the labels.  This noise transition matrix is then learned as a non-linear layer on top of the baseline model $M$ to denoise predictions.  The layer is discarded during final inference since gold labels are used during test time and are assumed to no longer be noisy. 
(2) The \textit{Crowdlayer} also operates by modeling the error probability, but assumes the noise arises due to annotator error, so a noise transition matrix is created for each worker \citep{rodrigues18crowdlayer}.  Once again, this matrix is learned with gradient descent and removed for final inference.
(3) The \textit{Forward} correction method from \cite{patrini17forward} adopts a loss correction approach which modifies the training objective. Given $-\log p(\hat{y} = \Tilde{y}|x)$ as the original loss, Forward modifies this to become $-log \sum_{j=1}^c T_{ji} p(\hat{y} = {y}|x)$ where $c$ is the number of classes being predicted, and both $i$ and $j$ are used to index the number of classes.  Matrix $T$ is represented as a neural network that is learned jointly during pre-training. % , which is shown to be sufficiently powerful in their paper. % They show in their paper % in Theorem 3
% that given sufficiently many corrupted samples, neural networks are powerful enough to model $T$.
(4) Lastly, the \textit{Bootstrap} method proposed by \cite{reed14bootstrap} generates pseudo-labels by gradually interpolating the predicted label $\hat{y}$ with the given noisy label $\Tilde{y}$.  We apply their recommended \textit{hard} bootstrap variant which uses the one-hot prediction for interpolation  since this was shown to work better in their experiments. % \footnote{We select the Bootstrap rather than CleanNet as our pseudo-label baseline since CleanNet becomes intractable with large language models, taking many months to run.} 

\subsection{Cleaning through Targeted Relabeling}
Rather than maximizing the number of examples annotated given our budget, we propose reserving a portion of the budget for reannotating the labels most likely to be incorrect.  Specifically, we start by annotating a large number of examples one time each using the majority of the budget $(n_a = 10,000; r=1)$.  We then pretrain a model $M_1$ using this noisy data, and observe either the model's training dynamics or output predictions to target examples for relabeling.  Next, we use the remaining budget to annotate those examples two more times $(n_b = 1,000; r=2)$, allowing us to obtain a majority vote on those examples. The final training set is formed by combining the 1k multiply-annotated examples with the remaining 9k singly-annotated examples.  We wrap up by initializing a new model $M_2$ with the weights from $M_1$ and fine-tune it with the clean data until convergence. % (See Figure 2 for an illustrative breakdown.) ---> TODO: design Figure 2
We experiment with four approaches for discovering the most probable noisy labels: 
\paragraph{Area Under the Margin} AUM identifies problematic labels by tracking the margin between the likelihood assigned to the target label class and the likelihood of the next highest class as training progresses \citep{pleiss20aum}. Intuitively, if the gap between these two likelihoods is large, then the model is confident of its argmax prediction, presumably because the training label is correct.  On the other hand, if the gap between them is small, or even negative, then the model is uncertain of its prediction, presumably because the label is noisy.  AUM averages the margins over all training epochs and targets the examples with the smallest margins for relabeling.

\paragraph{Cartography} Dataset Cartography is a technique for mapping the training dynamics of a dataset to diagnose its issues \citep{swayamdipta20carto}. The intuition is largely the same as AUM, such that Cartography also chooses consistently low-confidence (ie. low probability) examples for relabeling.  We take the suggestion from Section 5 of their paper to detect mislabeled examples by tracking the mean model probability of the true label across epochs.  Note that unlike AUM, Cartography tracks the final model outputs after the softmax, rather than the logits before the softmax. These can lead to different rankings since Cartography does not take the other probabilities in the distribution into account.

\paragraph{Large Loss} % \cite{zhang17understand} showed that over-parameterized deep neural networks are capable of fitting any dataset, and do so by first learning to fit to the clean data before memorizing the noisy data.  Building upon this, 
\cite{arpit17smallloss} found that correctly labeled examples are easier for a model to learn, and thus incur a small loss during training, whereas incorrectly labeled examples produce a large loss. Inspired by this observation and other similar works \citep{jiang18mentornet}, the Large Loss method selects examples for cleaning by ranking the top $n_b$ examples where the model achieves the largest loss during the optimal stopping point.  The ideal stopping point is the moment after the model has learned to fit the clean data, but before it has started to memorize the noisy data \citep{zhang17understand}.  We approximate this stopping point by performing early stopping during training when the progression of the development set fails to improve for three epochs in a row.  We then use the earlier checkpoint for identifying errors.

\paragraph{Prototype} We lastly consider identifying noisy labels as those which are farthest away compared to the other training data \cite{lee18cleannet}.  More specifically, we use a pretrained model to map all training examples into the same embedding space. Then, we select the vectors for each label class to form clusters where the centroid of each cluster is the ``prototype'' \cite{snell2017proto}.  Finally, we define outliers as those far away from the centroid for their given class, as measured by Euclidean distance, which are then selected for cleaning.

\section{Experiments}
% This section discusses experimental setup including pre-processing steps and model configurations.

\subsection{Datasets and Tasks} 
To test our proposal, we select datasets that span across four natural language processing tasks.  We choose these datasets because they provide multiple labels per example, allowing us to simulate single- and multiple-annotation scenarios.  % specific

% Task Descriptions
\paragraph{Offense} The Social Bias Frames dataset collects instances of biases and implied stereotypes found in text \citep{sap20offense}. We extract just the label of whether a statement is offensive for binary classification.

\paragraph{NLI} We adopt the MultiNLI dataset for natural language inference \citep{williams2018mnli}.  The three possible label classes for each sentence pair are \textit{entailment}, \textit{contradiction}, and \textit{neutral}.

\paragraph{Sentiment} Our third task uses the first round of the DynaSent corpus for four-way sentiment analysis \citep{potts20sentiment}.  The possible labels are \textit{positive}, \textit{negative}, \textit{neutral}, and \textit{mixed}.

\paragraph{QA} Our final task is question answering with examples coming from the NewsQA dataset \citep{trischler17newsqa}. The input includes a premise taken from a news article, along with a query related to the topic. The target label consists of two indexes representing the start and end locations within the article that extract a span of text answering the query.  Unlike the other tasks, the format for QA 
% QA is decidedly unique compared to the prior tasks in that the format 
is span selection rather than classification.  Due to this distinction, certain denoising methods that assume a fixed set of candidate labels are omitted from comparison.  %  on the QA task. A prediction is deemed correct when the predicted span of text matches the label, not accounting for leading or trailing spaces.

% Preprocessing methodology, budget-constrained experimental setup
\subsection{Training Configuration}
In our experiments, we fine-tune parameters during initial training with only six runs, which is composed of three learning rates and two levels of dropout at 0.1 and 0.05. Occasionally, when varying dropout had no effect, we consider doubling the batch size instead from 16 to 32. % This is to make a fair comparison across methods rather than relying on hyperparameter tuning to seek out the greatest gains.
We found an appropriate range of learning rates by initially conducting some sanity checks on a sub-sample of development data for each task and model combination. Learning rates were chosen from the set of [1e-6, 3e-6, 1e-5, 3e-5, 1e-4].  When a technique contained method-specific variables, we defaulted to the suggestions offered in their respective papers. We do not expect any of the methods to be particularly sensitive to specific hyperparameters.  % Note that we only use a subsample of the datasets, so we would not reach SOTA levels anyway. 

\subsection{Model Variations}
We select three models for comparison that represent strong options at their respective model sizes.  We repeat the process of example identification and simulated re-annotation separately for each model.  % training capacity.  
We use all models as a pre-trained encoders to embed the text inputs of the different tasks we study. 

DeBERTa-XLarge is our large model, which contains 750 million parameters and currently is the state-of-the-art on many natural language understanding tasks \citep{he21deberta}. % improves upon previous state-of-the-art Transformers by adding a novel disentangled attention mechanism and an enhanced mask decoder, both of which take better advantage of a token's position within a piece of text.  
DistilRoBERTa represents a distilled version of RoBERTa-base \citep{liu19roberta}. It contains 82 million parameters, compared to the 125 million parameters found in RoBERTa. Learning follows the distillation process set by DistillBERT where a student model is trained to match the soft target probabilities produced by the larger teacher model \citep{sanh19distill}.  Fine-tuning DistilRoBERTa is approximately 60-70 times faster compared to fine-tuning DeBERTa-XLarge on the same task.

\let\under\underline

\begin{table*}[ht]
\small \centering
    \subfloat[offense][Offensive Language Detection from SBF]{
        \resizebox{0.48\linewidth}{!}{ 
        \begin{tabular}{lcccc}
        \toprule
        \textbf{Methods} & \textbf{FastT} & \textbf{DRoB} & \textbf{DeXL} & \textbf{Avg} \\
        \midrule
        Oracle      & \textit{78.0} & \textit{81.8} & \textit{86.2} & \textit{82.0} \\
        Control     & 77.0          & 81.4          & 86.0          & 81.5          \\
        Majority    & 76.2          & 80.4          & 84.5          & 80.4          \\
        \midrule
        Adaptation  & \textbf{77.8} & 81.5          & \textbf{86.1} & \textbf{81.8} \\
        Crowdlayer  & 77.1          & 81.4          & 85.4          & 81.3          \\
        Bootstrap   & 77.1          & 81.2          & 85.1          & 81.2          \\
        Forward     & 77.5          & 81.2          & 84.9          & 81.2          \\
        \midrule
        Large Loss  & 77.7          & \textbf{81.6} & 85.4          & 81.6          \\
        AUM         & 77.5          & 81.5          & 85.3          & 81.4          \\
        Cartography & 77.3          & 81.2          & 85.0          & 81.2          \\
        Prototype   & 77.7          & 81.4          & 85.5          & 81.5          \\
        \bottomrule
        \end{tabular}}
        \label{tab:offense}}
    \
    \subfloat[nli][Natural Language Inference from MNLI]{
        \resizebox{0.48\linewidth}{!}{
        \begin{tabular}{lcccc}
        \toprule
        \textbf{Methods} & \textbf{FastT} & \textbf{DRoB} & \textbf{DeXL} & \textbf{Avg} \\
        \midrule
        Oracle      & \textit{40.7} & \textit{49.7} & \textit{88.3} & \textit{59.6} \\
        Control     & 40.1          & 48.5          & 87.4          & 58.7          \\
        Majority    & 38.5          & 46.2          & 86.1          & 56.9          \\
        \midrule
        Adaptation  & 40.6          & \textbf{49.4} & 87.8          & \textbf{59.2} \\
        Crowdlayer  & 40.2          & 48.7          & 87.4          & 58.7          \\
        Bootstrap   & \textbf{40.8} & 49.3          & 87.4          & 59.1          \\
        Forward     & 40.6          & 48.6          & 87.3          & 58.8          \\
        \midrule
        Large Loss  & 40.5          & 48.9          & 87.8          & 59.1          \\
        AUM         & 40.3          & 49.0          & 87.1          & 58.8          \\
        Cartography & 40.1          & 48.1          & 87.0          & 58.4          \\
        Prototype   & 40.4          & 48.6          & \textbf{88.0} & 59.0          \\
        \bottomrule
        \end{tabular}}
        \label{tab:nli}}

    \subfloat[sentiment][Sentiment Analysis from DynaSent]{
        \resizebox{0.48\linewidth}{!}{
        \begin{tabular}{lcccc}
        \toprule
        \textbf{Methods} & \textbf{FastT} & \textbf{DRoB} & \textbf{DeXL} & \textbf{Avg} \\
        \midrule
        Oracle      & \textit{55.5} & \textit{57.3} & \textit{73.2} & \textit{62.0} \\
        Control     & 54.0          & 57.2          & 72.7          & 61.3          \\
        Majority    & 52.4          & 55.8          & 71.2          & 59.8          \\
        \midrule
        Adaptation  & 53.8          & 56.8          & 72.6          & 61.1          \\
        Crowdlayer  & 53.9          & 57.2          & 72.7          & 61.2          \\
        Bootstrap   & 54.1          & \textbf{57.4} & 72.7          & 61.4          \\
        Forward     & 53.5          & 57.3          & 73.0   & 61.4          \\
        \midrule
        Large Loss  & \textbf{55.6} & \textbf{57.4} & \textbf{73.1} & \textbf{62.0} \\
        AUM         & 55.4          & 56.5          & 72.6          & 61.5          \\
        Cartography & 55.0          & 56.6          & 72.0          & 61.2          \\
        Prototype   & 55.1          & 57.1          & \textbf{73.1} & 61.7          \\
        \bottomrule
        \end{tabular}}
        \label{tab:sentiment}}
    \
    \subfloat[qa][Question Answering from NewsQA]{
        \resizebox{0.48\linewidth}{!}{
        \begin{tabular}{lcccc}
        \toprule
        \textbf{Methods} & \textbf{FastT} & \textbf{DRoB} & \textbf{DeXL} & \textbf{Avg} \\
        \midrule
        Oracle      & --- & \textit{7.94} & \textit{52.3} & \textit{30.1} \\
        Control     & --- & 6.90          & 50.3          & 28.6          \\
        Majority    & --- & 5.89          & 47.9          & 26.9          \\
        \midrule
        Adaptation  & --- &   ---         &   ---         &   ---         \\
        Crowdlayer  & --- &   ---         &   ---         &   ---         \\
        Bootstrap   & --- & 6.72          & 50.5          & 28.6          \\
        Forward     & --- &   ---         &   ---         &   ---         \\
        \midrule
        Large Loss  & --- & \textbf{6.95} & \textbf{51.5} & \textbf{29.2} \\
        AUM         & --- & 6.69          & \textbf{51.5} & 29.1          \\
        Cartography & --- & 6.24          & 51.0          & 28.6          \\
        Prototype   & --- &   ---         &   ---         &   ---         \\
        \bottomrule
        \end{tabular}}
        % \includegraphics[width=0.4\textwidth]{figure4.jpg}
        % \label{fig:subfig4}
        \label{tab:qa}}
\caption{Aggregated results for all method and model combinations, averaged over three seeds. Model names are abbreviated for space: FastT is FastText, DRoB is DistilRoBERTa, and DeXL is DeBERTa-XLarge.  Avg is the average across models for that method.  FastText doesn't produce context-dependent representations, and so is not usable on the QA task.}
\label{fig:results}
\end{table*}

For the final model, we avoid using Transformers altogether and instead use the FastText bag-of-words encoder \cite{joulin17fasttext}.  % to represent a really small model.
The FastText embeddings are left unchanged during training, so the only learned parameters are in the 2-layer MLP we use for producing the model's final output.  The same output prediction setup is used for all models, with a 300-dimensional hidden state.  Training the FastText models run roughly 100-120 faster compared to working with DeBERTa-XLarge.
\section{Major Results}
\label{sec:results}

Table \ref{fig:results} displays results across all models types and tasks, with each row representing a different technique.  All rows except the Oracle were trained using the same label budget of 12,000 annotations.\footnote{Our annotation amount is much less than total available data for a task so our results are not directly comparable to prior work.  For example, DynaSent train set includes 94,459 examples and Social Bias Frames contains 43,448 examples.} 
% The number of annotations we use is often much less than the total amount of available data for a task so our results are not directly comparable to prior state-of-the-art methods.  For example, the full DynaSent training set includes 94,459 examples and Social Bias Frames contains 43,448 examples.
In some cases, a method may surpass the Oracle since we conducted limited hyperparameter tuning,  %so the Oracle may do poorly on a particular seed, but 
but as expected, the Oracle model outperforms all other methods overall.  Notably, the Control setting always beats the Majority setting.  In fact, Majority is consistently the lowest-performing method on all models and tasks, showing that improved label quality is never quite enough to overcome the reduction in annotation quantity. %  redundant annotation is almost certainly never the best way to spend the budget.
Adaptation is the best among denoising methods, achieving the strongest results in two out of four settings.  Large Loss is the best among cleaning methods, with the highest scores in the remaining two tasks.  Prototypical is also a strong runner-up.  Large Loss is the best overall method due to its consistency since it never drops below second on all tasks.
%\vspace{-2cm}

Variance among the three seeds is fairly consistent for all models and methods within the same task.  Specifically, the standard deviation for offense detection and NLI are both around 0.5, with sentiment analysis and QA around 1.5 and 4.5, respectively.  We do not see any strong trends across tasks, nor any outliers for a specific method.  % particular 

\paragraph{Breakdown by Task}
Table 1a contains the results for offense language detection, where we see that Large Loss and Adaptation are the only methods to overtake the Control.  These two are also the best overall performers on natural language inference as seen in Table 1b.  The cleaning methods really shine on sentiment analysis and question answering where even the worst cleaning method often tops the best denoising method. We hypothesize this happens because the denoising methods work best in simple classification tasks, which we further explore in the next section.  A handful of results are not reported in Table 1d since they refer to methods that are designed exclusively for classification tasks, and cannot be directly transferred to span selection.

% \paragraph{Breakdown by Method}
%  and a close second in the final task.  AUM and Cartography are not great, but Crowdlayer and Forward generally perform even worse. 

\begin{table}[t]
\centering \includegraphics[width=0.5\textwidth]{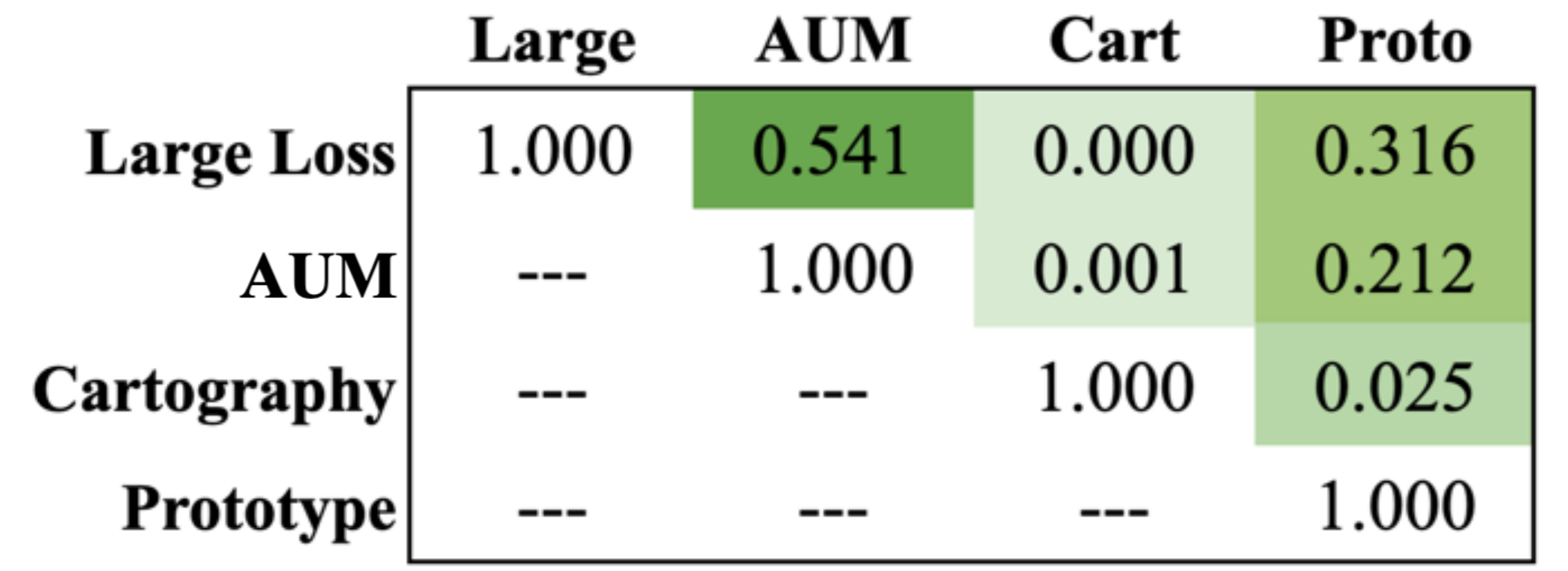}
\caption{Jaccard similarity for all pairs of targeted relabeling methods on the sentiment analysis task. Large, Cart and Proto are short for Large Loss, Cartography and Prototype, respectively.  Results for other tasks available in Appendix \ref{sec:quantitative}.}
\label{fig:sim_sa}
\end{table}

\paragraph{Breakdown by Model}
The larger models perform better than the smaller models in terms of downstream accuracy, but somewhat surprisingly, there does not seem to be any clear patterns in relation to the method.  In other words, if a particular method performs well (poorly) with one model size, it tends to also do well (poorly) when applied to the other model sizes too. One possible exception to this is the Prototype method showing strong performance with DeBERTA-XLarge. This is possibly because a stronger model produces more valuable hidden state representations for determining outliers.  Since method performance is largely independent of the model size, we use DistillRoBERTa as the encoder for simplicity in the upcoming analyses.

\begin{table}[t]
\small
\centering
%\resizebox{\linewidth}{!}{
{\begin{tabular}{lcccc}
    \toprule
    \textbf{Methods} & \textbf{Offense} & \textbf{NLI} & \textbf{Sentiment} & \textbf{QA} \\
    \midrule
    Default          & 81.6             & 48.9         & 57.4             & 6.95   \\
    Random           & 80.9             & 48.0         & 55.8             & 6.41   \\
    Cross            & 81.7             & 48.4         & 57.3             & 6.56   \\
    \bottomrule
\end{tabular}}
\caption{Ablation results that vary the method of identifying errors for relabeling. Default uses the same model for error selection and training.}  % , while Cross uses different models
\label{tab:ablation}
\end{table}

\paragraph{Ablation} How can we be sure that the cleaning methods are actually exhibiting a small, but consistent gain over the baselines rather than just natural variation? Perhaps the scores are close simply because all the methods use the same amount of training data. If the cleaning methods are indeed adding value, then they should perform much better than random selection.
To measure this, we replace the pre-trained DistilRoBERTa model with a uniform sampler to identify examples for cleaning. 

Active learning has been shown to exhibit significant decrease when transferring across model types \citep{lowell19activelearn}.  In contrast, we argue that our method is not active learning since it is not directly dependent on the specific abilities of the target model.  To test this claim, we also conduct an additional ablation whereby we replace one model type for another. Namely, we use the DeBERTa-XLarge model to select examples for cleaning, then train on the DistilRoBERTa model. %  In both cases, we then train a DistilRoBERTa model from scratch as usual.

The results in Table~\ref{tab:ablation} show that randomly selecting data points to relabel indeed lowers the final performance by a noticeable amount.  By comparison, cross training models 
% using data cleaned by another model type 
leads to a negligible drop in performance. We believe this occurs because targeted relabeling produces clean data, and clean data is useful regardless of the situation. % setup, whereas active learning chooses examples tailored for specific s.

\section{Discussion and Analysis}
To better understand how the proposed clean methods operate, we conduct additional analysis with the sentiment analysis task.% For the sake of distinguishing between the two sets of methods, we refer to the four noise correction baselines as denoising methods and our proposed techniques as cleaning methods.

\begin{table}[ht]
\small
\centering
%\resizebox{\linewidth}{!}{
{\begin{tabular}{lccc}
    \toprule
    \textbf{Methods} & \textbf{Precision} & \textbf{GoEmotions} & \textbf{Synthetic} \\
    \midrule
    Oracle           & ---        & \textit{55.8} & \textit{57.9} \\
    Control          & ---        & 54.8          & 56.6          \\
    Majority         & ---        & 53.0          & 55.2          \\
    \midrule
    Adaptation       & ---        & 54.8          & 56.5          \\
    Crowdlayer       & ---        & 54.9          & 56.4          \\
    Bootstrap        & ---        & 55.0          & \textbf{57.0} \\
    Forward          & ---        & 53.9          & 56.2          \\
    \midrule
    Large Loss       & 56.8       & \textbf{55.2} & 56.5          \\
    AUM              & 60.4       & 54.6          & 56.1          \\
    Cartography      & 19.0       & 54.3          & 56.4          \\
    Prototype        & 46.6       & 55.1          & 56.7          \\
    \bottomrule
\end{tabular}}
\caption{This table contains results for the three different post-hoc analyses.  Left column is precision of the model in identifying mislabeled examples. Right columns are results training on extended datasets. All scores are average of three seeds on DistillRoBERTa.}
\label{tab:extended_datasets}
\end{table}

% Part 1: Why does clean methods work so well? Compare clean to advanced.
% \begin{enumerate}
%     \item They assume many annotations per example - unrealistic in real-life settings
%     \item They use vision rather than text data, but text data is more nuanced.  Even wrong answers can be partially correct.   NLP due to difficulty from inherent ambiguity \cite{pavlick19inherent, chen20uncertain}  We are moving from images to text.  More specifically text noise is often not so obvious, mistakes are more subtle.
%     \item They use artificially generated errors - explains why the noise transition matrix items are poor, synthesized label noise
%     \item Many other papers tackle under "high noise" perturbations with 50\% or more of the labels being corrupted.  High noise ratio, high proportion of wrong labels.
%     \item they make assumption that the crowdsource label is independent of the data given the ground truth label, but this is patently false.
% \end{enumerate}

\paragraph{How well do clean methods select items?}
We compare the four proposed methods by first looking at the amount of overlap in the examples selected for relabeling.  To calculate this, we gather all examples chosen for relabeling by their likelihood of annotation error. For a given pair of methods, we then find the size of their intersection and divide by the size of their union, which yields the Jaccard similarity.  As shown in Table \ref{fig:sim_sa}, AUM and Large Loss have high overlap meaning that they select similar examples for cleaning. % while Cartography has low overlap with everything else. Prototype lies somewhere in the middle. 
We additionally calculate the precision of each method by counting the number of times a label targeted for relabeling did not match the oracle label, and therefore legitimately requires cleaning.  Based on Table \ref{tab:extended_datasets}, we once again see reasonable performance for the Large Loss cleaning method. % that Cartography performs abnormally, while the other methods are roughly equal to each other.  % All this lines up nicely with downstream model performance, implying that the ability of the cleaning methods to identify mislabeled examples directly contributes to their level of impact.

\begin{table*}[ht]
\small
\centering
{% \resizebox{\pagewidth}{!}{
% \begin{tabular}{l|c>{\centering}p{2.9em}c}
\begin{tabular}[width=\textwidth]{lp{38em}c}
% \Xhline{1.4pt}
\toprule
\textbf{Method} & \textbf{Input Text} & \textbf{Label}\\
% \hhline{|=|====|}       \hdashline
\midrule
        & That's usually how it go goes.                                & \textsc{mixed}    \\
        & I always order ``to-go''                                      & \textsc{mixed}    \\
\textbf{Large Loss}  
        & It's \$15 bucks for a beer since I used a drink ticket        & \textsc{mixed}    \\
        & We usually frequent the settlers ridge location.              & \textsc{mixed}    \\
        & I went on June 4th around 10:30.                              & \textsc{mixed}    \\
\midrule
        & So fine, no problem.                                          & \textsc{positive} \\
        & A sirloin hotdog wrapped in bacon.                            & \textsc{neutral}  \\
\textbf{AUM}     
        & For many years, I have gone to the Pet Smart down the street. & \textsc{neutral}  \\
        & I was always so happy here when it was managed by Johnny.     & \textsc{neutral}  \\
        & I ordered the pad Thai noodles, chicken chow mien and egg rolls.    & \textsc{positive} \\
\hline
        & The food and customer service was fantastic when you first opened   & \textsc{positive} \\
        & The servers were pleasant.                                    & \textsc{positive} \\
\textbf{Cartography}
        & Our waiter was overly friendly and informational.             & \textsc{mixed}    \\
        & Family owned and operated these folks are killing it          & \textsc{positive} \\
        & I really thought the young folks behind the counter were outgoing and seemed to enjoy their jobs
        & \textsc{positive}  \\  
\midrule
        & This should be a fun family place!                            & \textsc{negative} \\
        & Hotel was awesome.                                            & \textsc{negative} \\
\textbf{Prototype}
        & Great service for many years on our cars, but always at an additional price. & \textsc{neutral} \\
        & Salad was great but a bit small.                              & \textsc{neutral}  \\
        & We had to specify the order \textit{multiple} times, but eventually when the food came it was actually pretty good.
        & \textsc{neutral} \\  
\bottomrule
\end{tabular}}
\caption{Sentiment Analysis examples each method identified as being most likely to be label errors.} \label{tab:rank_sent}
\end{table*}

% Prior papers make the assumption that the crowdsource label is independent of the data given the ground truth label, but this is patently false.
%  Dawid \& Skene (1979), Raykar et al. (2010),  Albarqouni et al. (2016),  Guan et al. (2017) , Rodrigues \& Pereira (2017), Max-MIG (Cao et al. 2020)
% Authors claim that "no algorithm can train a classifier here to avoid the influence of the image information" when these items are dependent, for example taking into account the "indoor or outdoor'' aspect of the image when classifying dogs vs. cats. "we do not consider the image difficulty", but believe mixing these ideas would be beneficial for future work
% This is totally true!  Which is why our solution is not to apply an advanced classifier, but instead to have extra human annotators provide redundant annotations.  Rather than relying on some classifier, we rely on human ability.
% papers such as from  \citet{khetan16achieve} go beyond restricting workers to have a common confusion matrix for all questions. In this respect, these are better aligned with the realistic scenario where the error in labeling may depend on the closeness to the decision boundary.  \cite{shah21permutation}

Qualitative examples for sentiment analysis are displayed in Table \ref{tab:rank_sent}, which were chosen as the most likely examples of label errors according to their respective methods. Large Loss consistently discovers `neutral' labels that were mis-labeled as `mixed', while Prototype also does a good job uncovering label errors, finding `positive' examples mislabeled as `negative'.  Overall, we see that the best performing cleaning methods do seem to pick up on meaningful patterns.

% and `mixed' examples mislabeled as `neutral'.  Cartography seems to have picked up on more nuanced examples that may be harder to learn.  For example, `The servers were pleasant.' sounds positive, but could also be seen as negative if the statement is interpreted to mean that the servers were previously pleasant, but no longer are any more.   % Cartography is less reliable in selecting labeling errors, which helps to explain its lower quantitative performance. 

% explain, correlates, align, matches our expectation

\begin{figure}
  \includegraphics[width=\linewidth]{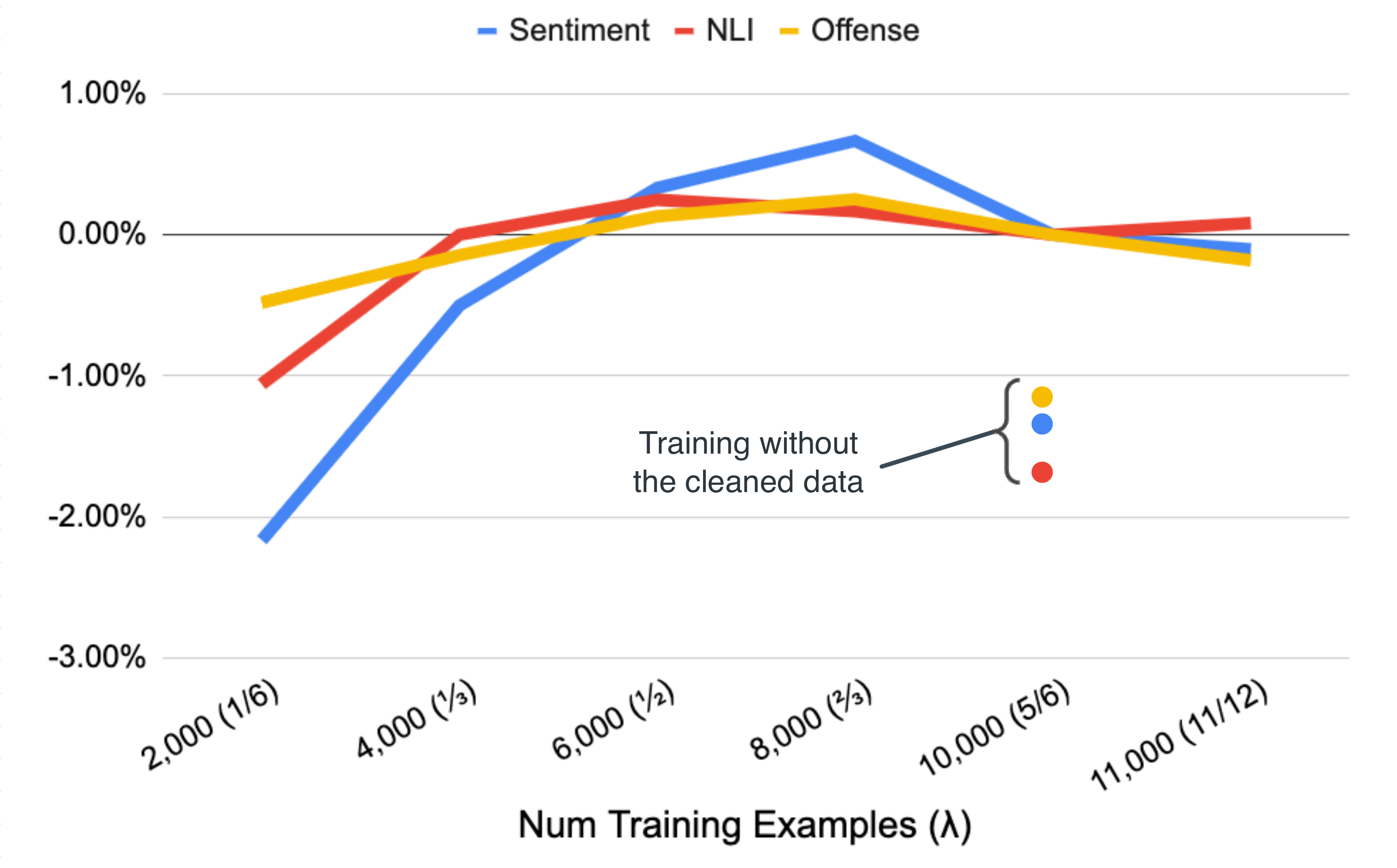}
  \caption{Varying the number of training examples changes the amount of budget remaining for cleaning.  10,000 examples is set as the default and the percent change is measured in comparison to this point.}
  \label{fig:lambda}
\end{figure}

\paragraph{How many examples should be cleaned?}
All cleaning experiments so far have been run with $n_a$ = 10,000 examples with $n_b$ = 1,000 samples chosen for relabeling.  This is equivalent to using up $\lambda = \frac{5}{6}$ of the labeling budget upfront, with the remaining annotations saved for later.  This $\lambda$ ratio was chosen as a reasonable default, but can also be tuned to be higher or lower.  Figure~\ref{fig:lambda} shows the results of varying the $\lambda$ parameter from a range of $\frac{1}{6}$ to $\frac{11}{12}$. Based on the results, choosing $\lambda = \frac{2}{3}$ would have actually been the best option.  This translates to $n_a$ = 8,000 examples with $n_b$ = 2,000 of those examples selected for re-labeling.  As a sanity check, we also try dropping the $n_b$ cleaned examples when retraining, keeping only the noisy data.  As seen in Figure~\ref{fig:lambda}, the performance decreases as expected.

\paragraph{What if we increase the number of classes?}
Based on the trends in the task breakdown of section \ref{sec:results}, denoising methods seem to perform worse than explicit relabeling methods as the task gets harder.  Most denoising methods may even become intractable for complex settings, such as those which require span selection. % question answering and named entity recognition that require span selection.
To test this hypothesis, we extend our setup to the GoEmotions dataset, where the goal of the task is to predict the emotion associated with a given utterance % containing 28 classes
\cite{demszky2020goemotions}. Whereas previous tasks dealt with 2-4 classes, the GoEmotions dataset requires a model to select from 27 granular emotions and a neutral option, for a total of 28 classes.  Intuitively, we would expect the denoising methods to struggle since the pairwise interactions among classes has grown exponentially larger.  The results in Table \ref{tab:extended_datasets} reveal that Large Loss again outperforms all other methods in prediction accuracy. Notably, Adaptation in particular continues to exhibit lower than average scores compared to other methods.  This supports our claim that relabeling methods are more robust as the number of classes grows.

\paragraph{What happens if noise is synthetically created?}
Many of the advanced denoising methods were originally tested on synthetically generated noise, whereas the noise in our datasets originates from noisy annotations, caused by the inherent ambiguity of natural language text \cite{pavlick19inherent, chen20uncertain}. Perhaps this partially explains how our proposed relabeling methods are able to outperform prior work.  To study this further, we create a perturbed dataset starting from the gold DynaSent examples.  Specifically, we randomly sample replacement labels according to a fabricated noise transition matrix, rather than sampling from the label distribution provided by annotators.  (More details in Appendix \ref{sec:preprocess}.)  With noise coming from an explicit transition matrix, it might be easier for all models to pick up on this pattern.

The middle column of Table \ref{tab:extended_datasets} shows that all eight cleaning methods perform on par with each other.  When comparing the variance on this dataset with synthetic noise against the original DynaSent dataset with natural noise, the standard deviation drops from 0.34 down to 0.28, highlighting the uniformity in performance among the eight methods.  The denoising methods work as intended on synthetic noise, but such assumptions may not hold on real data with more nuanced errors. 
% This leaves relabeling as the reliable methid which performs reliably in practice.
%while the denoising methods fail to significantly outperform the relabeling methods as expected, Large Loss has successfully been dethroned as the top ranking method.  Rather, 

\section{Conclusion}
Noisy data is a common problem when annotating data under low resource settings.  Performing redundant annotation on the same examples to mitigate noise leads to having even less data to work with, so we propose data cleaning instead through targeted relabeling. % study of learning with noisy data in practical settings.
We apply our methods on multiple model sizes and NLP tasks of varying difficulty, which show that
% Our comparisons 
saving a portion of a labeling budget for re-annotation matches or outperforms other baselines despite
% for learning with noisy data. Our strongest method, Large Loss, not only achieves high model accuracy, but is also one of the easiest techniques to implement, 
requiring no extra parameters to train or hyper-parameters to tune. Intuitively, our best method can be summarized as double-checking the examples that the model gets wrong to see if it is actually an incorrect label causing problems.

Thus, to make the most out of the scarce labeled data available, we believe a best practice should include targeting examples for cleaning rather than spending the entire annotation budget upfront.  
Future work includes exploring more sophisticated techniques for identifying examples to relabel and understanding how such cleaning models perform on additional NLP tasks such as machine translation or dialogue state tracking, which have distinct output formats.

% our models work better because (1) we don't make so many false assumptions (2) they are not so complicated with many moving parts in code and hyper-param tuning

% We hope others will follow by considering data cleaning as a baseline when trying to annotate data with a fixed budget.
% clean based on some insight: easy examples, redundant examples, noisy examples, already known, possibly distracting, too hard

\section*{Acknowledgements}
The authors would like to sincerely thank the reviewers for their attention to detail when reading through the paper.  Their insightful questions and advice have noticeably improved the final manuscript.  We would also like to thank ASAPP for sponsoring the costs of this project.  Finally, we would like to acknowledge the helpful feedback from discussions with members of the Columbia Dialogue Lab.

\clearpage
\bibliography{custom}
\bibliographystyle{acl_natbib}

\clearpage
\appendix

\begin{figure*}[ht]
\centering
    \subfloat[sim_od][Jaccard similarity on Social Bias Frames]{
        \includegraphics[width=0.5\textwidth]{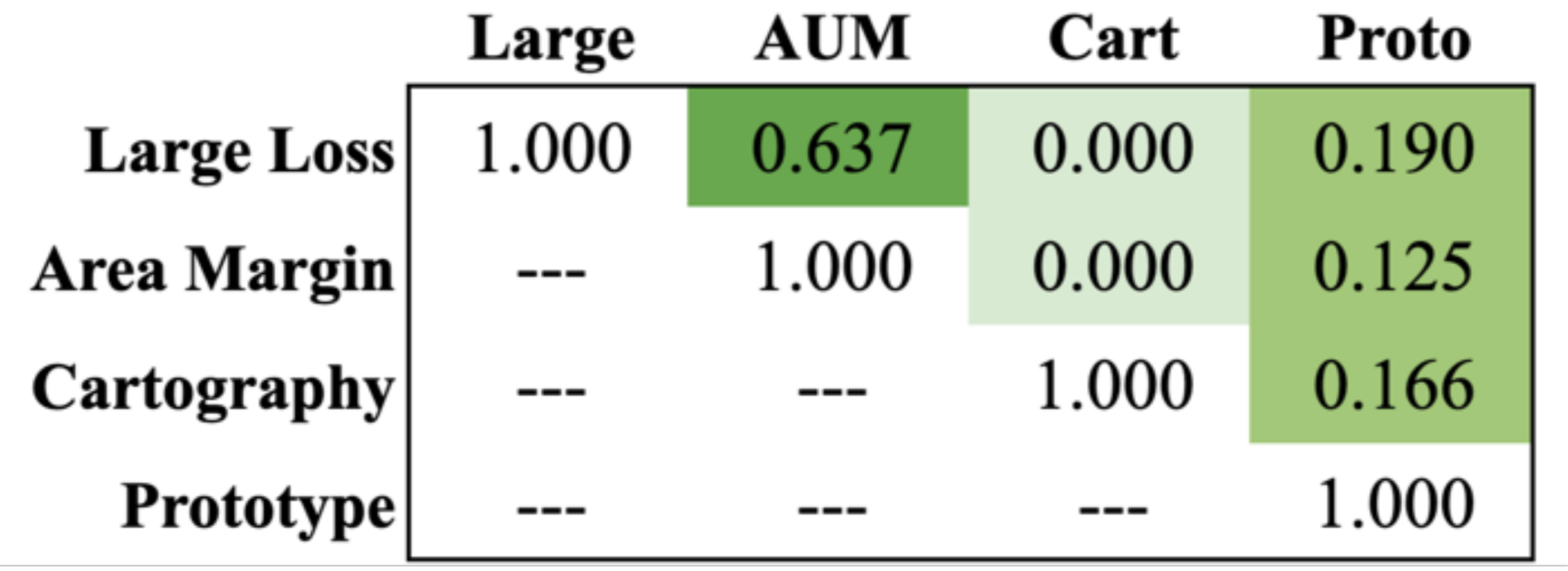}
        \label{fig:sim_od}}
    \subfloat[sim_nli][Jaccard similarity on MNLI dataset]{
        \includegraphics[width=0.5\textwidth]{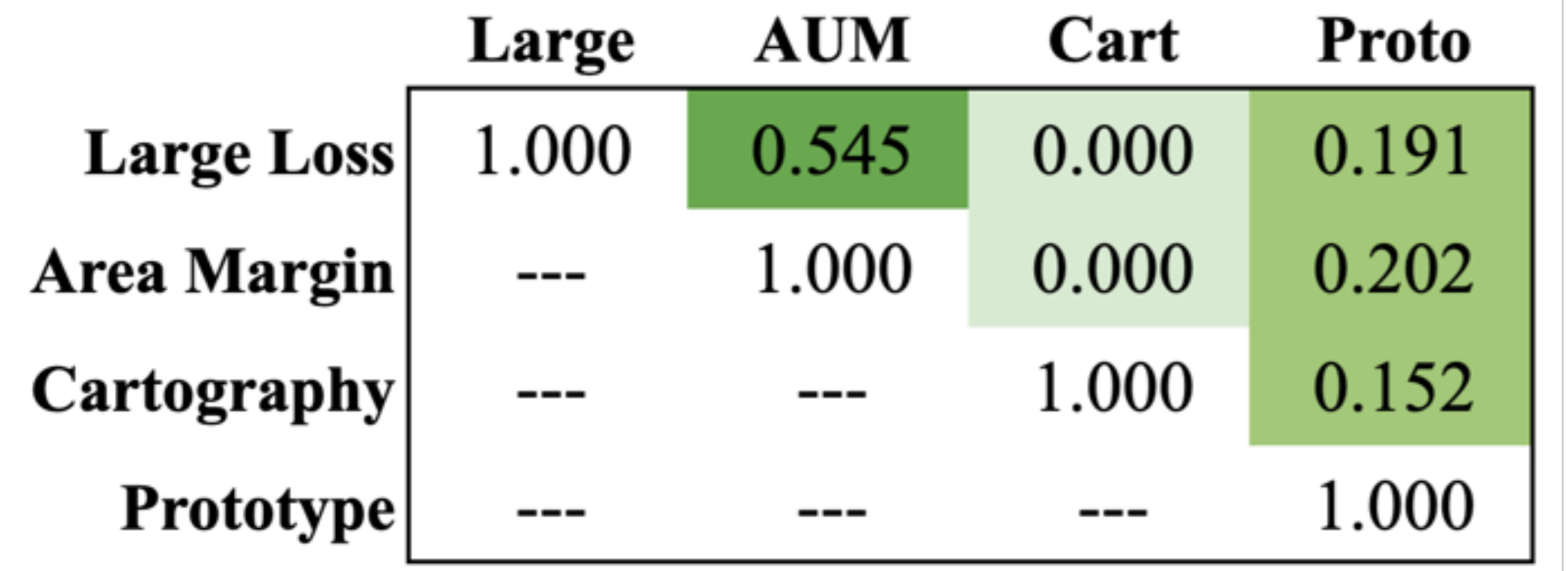}
        \label{fig:sim_nli}}
\caption{Jaccard similarity overlap for all pairs of targeted relabeling methods on the offensive language detection task and the natural language inference task.}
\label{fig:jaccard}
\end{figure*}

\section{Additional Quantitative Results}
\label{sec:quantitative}
Looking at Figure~\ref{fig:jaccard}, the similarity scores for offensive language detection and natural language inference largely match up with the scores found in sentiment analysis.  In particular, Large Loss and AUM exhibit higher overlap with each other.  Additionally, Prototype shows a medium overlap and Cartography shows no overlap at all with the other methods. We reach a similar conclusion that the Large Loss method is a reasonable technique.

\section{Additional Qualitative Examples}
\label{sec:qualitative}
More examples can be found in Table~\ref{tab:rank_nli} on the next page.  We see that Large Loss is once again quite accurate in picking up labeling errors.  Prototype for NLI does a great job at finding examples labeled as `entailment' which should be something else.  The hypotheses for all the selected examples contain negative sentiment, which may be located far away from the entailment examples in the embedding space.  Cartography exhibits a pattern of always choosing examples labeled as `contradiction'. % However, once again Cartography have selected examples there were correctly labeled and did not need additional annotation.  This aligns with the low precision scores in Table \ref{tab:extended_datasets}.

\section{Comparison to Learning Schemes}
\label{sec:scheme}
On the surface, targeting examples for relabeling contains may seem similar to active learning or curriculum learning.  Although there are certainly some parallels between these techniques, these are fundamentally different learning paradigms.

Active learning methods typically choose new examples to label based on the uncertainty of the model~\cite{tong01svm, hanneke14active} or on the diversity they can add to the existing distribution~\cite{sener18coreset, ash20badge}.  Although sample noise can also be measured through model uncertainty, denoising and active learning do not have the same goal.  More specifically, the noise of a training example is related to how its label is somehow incorrect.  Perhaps the start of a span was not properly selected or an example that should not be tagged was accidentally included.  More simply, an example is mislabeled as class A, when in fact it belongs to class B.  This situation is not possible with active learning because the examples in active learning do not have labels yet!  The entire point of active learning is to choose which examples should be labeled next~\cite{settles2008active, settles2011closing}.  Thus, when we try to identify examples for cleaning, we are \textit{re}-labeling rather than labeling for the first time.

Curriculum learning also selects examples for training based on model uncertainty~\cite{bengio09curriculum} and diversity maximization~\cite{jiang14diversity}.  It could be interpreted that easier examples are those that contain less noise, which would connect to our proposed process.  However, traditional curriculum learning chooses these examples upfront rather than based on modeling dynamics~\cite{jiang15selfpaced}.  Extensions have been made under the umbrella of self-paced curriculum learning whereby examples are chosen for a curriculum based on how they react to a model's behavior~\cite{kumar10selfpaced}.  This is indeed similar to how we can choose to relabel examples based on the model loss.  This aspect of relabeling though is the key distinction -- we \textit{act} on these examples in an attempt to denoise the dataset.  On the other hand, self-paced learning simply feeds those same examples back into the model without any modification.

\section{Data Preprocessing}
\label{sec:preprocess}

\subsection{Synthetic Data Generation}
The synthetic dataset is created by applying an explicit noise transition matrix with 20\% noise.  Since the original dataset contains four classes, we start with an empty 4x4 matrix.  The labels should not be confused most of the time so we assign a likelihood of 0.8 across the diagonal of the matrix.  Next, we randomly select another class for each row to receive 0.1 likelihood of confusion.  This leaves 0.1 for each row to be divided between the two remaining classes, which receive 0.05 each.  For each example in the oracle dataset, we use the original label to select a single row from the constructed noise transition matrix.  Lastly, we are able to sample a new label according to the weights provided by this 4-D vector.  In contrast, the original sampling procedure obtained its weights according to the normalized label distribution provided by the annotations.

\subsection{GoEmotions Preprocessing}
To prepare the GoEmotions dataset, we filter the raw data to include only examples that have at least three annotators and a clear majority vote (used for determining the gold label).  We then cross-reference this against the proposed data splits offered by the authors which have high inter-annotator agreement.  From this joint pool of examples, we sample 12k training examples to match the setting of all our other experiments.  This results in 12000/2954/2946 examples for train, development and test splits respectively.
 
\begin{table*}[ht]
\centering
\resizebox{\textwidth}{!}{
\begin{tabular}{rllc}
\toprule
\textbf{Method} & \textbf{Premise} & \textbf{Hypothesis} & \textbf{Label}\\
\midrule
           & Why shouldn't he be?                                                                       &   
             He doesn't actually want to be that way.                   & \textsc{entailment}    \\
           & How do they feel about your being a Theater major?         &
             They don't know you're a theater major, do they?           & \textsc{entailment}    \\
           & Defication of humankind as supreme.                                        & 
                         Humankind is not supreme.                      & \textsc{entailment}    \\
\textbf{Large Loss} & These are artists who are either emerging as leaders in their & 
            These artists are becoming well known in                & \textsc{contradiction} \\
           &  \quad fields or who have already become well known.   
           &  \quad  their fields.                                  &               \\
           & As he stepped across the threshold, Tommy brought      &
               Tommy stepped across a threshold and put             & \textsc{contradiction} \\
           & \quad the picture down with terrific force on his head. 
           & \quad  a picture  down on his head.                    &               \\
\midrule
           & And if, as ultimately happened, no settlement resulted,                                                                     
           & Even if an agreement could not be reach                & \textsc{entailment}    \\
           & \quad we could shrug our shoulders, say, 'Hey, we tried.'
           & \quad we could say we tried.                           &               \\
           & Companies that were foreign had to accept Indian
           & Foreign companies had to take Italian money            & \textsc{contradiction} \\
           & \quad financial participation and management.   &      &               \\
\textbf{AUM}    & ... he's been tireless about pursuing both celebrity   &
            He never wanted any attention and kept to               & \textsc{contradiction} \\
           & \quad and the cause of popular history ever since.   
           & \quad himself all the time.                                        &               \\
           & Two more weeks with my cute TV satellite dish  &
              My appreciation of my satellite dish has              & \textsc{entailment} \\
           & \quad  have increased my appreciation of it.        
           & \quad increased.                                       & \\
           & Each working group met several times to develop
           & Each working met more than once to discuss             & \textsc{entailment} \\
           & \quad  recommendations for ... legal services delivery system  
           & \quad changes to the legal services delivery system.           &            \\ 
\midrule
            & A detailed English explanation of the plot is always
            &    You'll have to figure the plot out on your own.        & \textsc{contradiction} \\
            & \quad provided, and wireless recorded commentary units ... &     &    \\
            & I just loved Cinderella . I also saw my sisters as the
            & I really disliked Cinderella and could never          & \textsc{contradiction} \\
            & \quad wicked stepsisters sometimes, and I was Cinderella ...
            & \quad relate to her.                                  &               \\
\textbf{Cartography} & The judge gave vent to a faint murmur of disapprobation & 
            The prisoner in the dock remained still and             & \textsc{contradiction} \\
            & \quad and the prisoner in the dock leant forward angrily.
            & \quad and expressionless                              &               \\
            & Jon was about to require a lot from her.
            & Jon needed nothing to do with her.                    & \textsc{contradiction} \\
            &  I know you'll enjoy being a part of the Herron School 
            &  You will detest the Herron School of Art and         & \textsc{contradiction} \\
            & \quad of Art and Gallery. 
            & \quad Gallery and have nothing to do with it          &               \\
\midrule
           & Why shouldn't he be?                                                                       &   
             He doesn't actually want to be that way.                   & \textsc{entailment} \\
           & I like this area a whole lot and it's, it's growing so much
           & I really despise living in this location and would     & \textsc{entailment} \\
           & \quad and I just want to be near my family ...
           & \quad prefer to be farther away from my relatives.     &            \\
\textbf{Prototype}  & The air is warm. & 
            The arid air permeates the surrounding land.            & \textsc{entailment} \\
           & Inside the Oval: White House Tapes From FDR to Clinton &
               No tapes were recorded in the white house            & \textsc{entailment} \\
           & He became even more concerned as its route changed     & 
           He wasn't worried at all for the plane                   & \textsc{entailment} \\
           & \quad moving into another sector's airspace.   &       &            \\ 
\bottomrule
\end{tabular}}
\caption{Natural language inference examples that each method identified as being most likely to be label errors.  Sentences were truncated in some cases for brevity.} \label{tab:rank_nli}
\end{table*}
 
 \section{Limitations}
Our proposed methods are limited to studying noise which comes from human annotators acting in good faith. Other sources of label noise include errors which occur due to spammers, distant supervision (as commonly seen in Named Entity Recognition), and/or pseudo-labels from bootstrapping.  Within interactive settings, such as for dialogue systems, models may also encounter noisy user inputs such as out-of-domain requests or ambiguous queries.  Our methods would not work well in those regimes either.
 
\end{document}